\documentclass[12pt,journal,draftcls,letterpaper,onecolumn]{IEEEtran}
\usepackage{cite}      % Written by Donald Arseneau
\usepackage{graphicx}  % Written by David Carlisle and Sebastian Rahtz
\usepackage{subfigure} % Written by Steven Douglas Cochran

\usepackage{amsmath}   % From the American Mathematical Society
\begin{document}
\title{Bayesian Nonlinear Principal Component Analysis Using Random Fields}
\author{Heng~Lian,~Nanyang Technological University
%\thanks{Manuscript received January 20, 2002; revised August 13, 2002.}% <-this % stops a space
\thanks{H. Lian is with Division of Mathematical Sciences, SPMS, Nanyang Technological University, Singapore. Email: henglian@ntu.edu.sg.}}
% note the % following the last \IEEEmembership and also the first \thanks -
% instance, "A\textbf{} \textbf{}B" will typeset as "A B" not "AB". If you want
% "AB" then you have to do: "A\textbf{}\textbf{}B"
\markboth{}{}

% If you want to put a publisher's ID mark on the page
% (can leave text blank if you just want to see how the
% text height on the first page will be reduced by IEEE)
%\pubid{0000--0000/00\$00.00~\copyright~2002 IEEE}

%\specialpapernotice{(Invited Paper)}

\maketitle

\begin{abstract}
We propose a novel model for nonlinear dimension reduction motivated by the probabilistic formulation of principal component analysis. Nonlinearity is achieved by specifying different transformation matrices at different locations of the latent space and smoothing the transformation using a Markov random field type prior. The computation is made feasible by the recent advances in sampling from von Mises-Fisher distributions.
\end{abstract}


\section{Introduction}
% \PARstart{A}{demo} file is ....
% \PARstart{A}{}demo file is ....
% \PARstart{T}{his demo} file is ....
\PARstart{P}{rincipal} component analysis (PCA) is an old statistical technique for unsupervised dimension reduction. It is often used for exploratory data analysis with the objective of understanding the structure of the data. PCA aims to represent the high dimensional data points with low-dimensional representers commonly called latent variables, which can be used for visualization, data compression etc. Sometimes PCA is also used as a preprocessing step before regression \cite{hastie01} or clustering \cite{liu03}. In these context, however, PCA typically does not have satisfying performance due to the ignorance of subsequent analysis.

We denote the original high dimensional data by $Y=\{y_1,y_2,\ldots y_n\}^T$, where $y_i=\{y_{i1},\ldots, y_{ip}\}^T\in R^p$. Note that the superscript $T$ is used to denote transposition so that $y_i$ is a column vector. We assume the data are already centered so that $\bar{y}=\sum_{i=1}^n y_i/n=0$.  One common definition of PCA is that of taking a linear combination of the components of $y_i$:
\[x_i=\sum_{j=1}^p y_{ij}v_j, i=1,2,\ldots, n\]
where $v_j$ is the weighting coefficient of the $j$-th covariate. This can be written as 
\begin{equation}\label{eqn:pca}
x_i=v^T y_i
\end{equation}
where $v=(v_1,\ldots,v_p)^T$. We take $||v||=1$ so that (\ref{eqn:pca}) represents a projection onto the linear subspace spanned by $v$. Given $v$ and $x_i$, the optimal linear reconstruction of $y_i$ is given by $\hat{y}_i=vx_i$.   We want $\hat{y}_i$ to be a good representation of the original $y_i$. Thus we aim to minimize $\sum_i||y_i-\hat{y}_i||^2=\sum||y_i-vx_i||^2$. It can be shown that the minimizing $v$ is the eigenvector of $Y^TY/n=\sum y_i^Ty_i/n$ associated with its largest eigenvalue, called the first principal component and denoted by $v_1$. Similarly, we can define $d (d\le p)$ principal components $v_1,\ldots,v_d$ as the minimizer with respect to $V$ of the total squared reconstruction error $\sum||y_i-\hat{y}_i||^2$, where $\hat{y}_i=Vx_i$, $V=(v_1,v_2,\ldots,v_d), V^TV=I_{d\times d}$, and $x_i=V^Ty_i\in R^d$ is the projection of $y_i$ onto the subspace spanned by the columns of $V$, the principal components.

PCA is a linear procedure since the reconstruction is based on a linear combination of the principal components. Several nonlinear extensions have been proposed. The most famous one in the statistical literature is the principal curves proposed in \cite{hastie89}. The principal curve is defined as the curve such that each point on the curve is the center of all the data points whose projection onto the curve is that point. Thus visually the principal curve is defined as the curve that passes through the ``middle" of the data points. Although conceptually appealing, the computational constraint makes it difficult to extend this approach to higher dimensions. Other approaches including neural networks \cite{kramer91}, kernel embedding \cite{webb96}, and generative topographic mapping \cite{bishop98} have been proposed.

The absence of probabilistic models in traditional PCA motivated the probabilistic PCA (PPCA) approach adopted by \cite{tippingpca99}. The advantage of probabilistic modeling is multifold, including providing a mechanism for density modeling, determination of degree of novelty of a new data point, and naturally incorporating incomplete observations. In \cite{tippingpca99}, the generative model is defined through the observation equation:
\begin{equation}\label{eqn:obs}
y_i=Wx_i+\epsilon_i
\end{equation}
which stated the linear relationship between the latent variable and the data points, $W$ is a $p\times d$ matrix that is not constrained to have orthogonal columns a priori, and $\epsilon_i$ are i.i.d. noises with $\epsilon_i\sim N(0,\sigma^2I_{p\times p})$. Note we assume that the data is already centered, otherwise the observation model should be changed to
\[y_i=Wx_i+\mu+\epsilon_i\]
with shift parameter $\mu$.
In PPCA, we put a zero mean, unit covariance Gaussian prior on $x_i$, and the likelihood is maximized over $(W,\sigma^2)$ after marginalizing over $x_i$:
\[\max \prod_i p(y_i|W,\sigma^2)=\max \prod_i\int p(y_i|x_i,W,\sigma^2)p(x_i)dx_i\] 
It is shown that when the noise level $\sigma$ goes to zero, the maximum likelihood estimator for $W$ will converge to 
\begin{equation}\label{eqn:mle}
W=VD,
\end{equation}
where the matrix $V$ and $D$ comes from singular value decomposition of $Y/\sqrt{n}=UDV^T$.
Thus PPCA is a natural extension of the traditional PCA.  

\cite{tipping99} extends PPCA to mixture PPCA which can be used to model nonlinear structure in the data. In PPCA, after marginalizing over $x_i$, the distribution of $y_i$ becomes $N(\mu,WW^T+\sigma^2I)$ if the data are not centered. The mixture PPCA models the marginal distribution of $y_i$ as 
\[p(y_i)=\sum_{m=1}^M \pi_mp(y_i|m),\]
a mixture with $M$ components, and for each component, the observation model is
\[y_i=W_mx_i+\mu_m+\epsilon_i\]
if the $i$-th observation comes from the $m$-th mixture component. Thus in mixture PPCA, each mixture component is defined by a different linear transformation, while clustering is defined on the original $p-$dimensional space. Marginalization over $x_i$ is still the same using unit covariance Gaussian distribution. The maximization over $\{W_m\}$ and $\{\mu_m\}$ can be performed using EM algorithm taking the mixture indicators as the missing data. The experiments in \cite{tipping99} showed that this model has a wide applicability. We also note that when using $x_i$ to reconstruct the data point $y_i$, we must also store the mixture component which is responsible for generating $x_i$, or, more preferably, the posterior responsibility of each mixture for the $i-$th observation. This piece of information cannot be recovered from the latent variable $x_i$ alone.

  Another approach of probabilistic nonlinear PCA based on Gaussian processes has been proposed in \cite{lawrence05}. It starts from the same observation model (\ref{eqn:obs}), but instead of marginalizing over $x_i$, it marginalizes over $W$ by putting independent spherical Gaussian prior on the $d$ columns of $W$, resulting in the marginal distribution of $y_{.j}\sim N(0,XX^T+\sigma^2I)$, where $y_{.j}$ is the $j$-th column of $Y$ and $X$ is the $n\times d$ matrix of latent variables. The author noticed that one can replace $ XX^T+\sigma^2I$ with another kernel matrix to achieve nonlinearity. Conceptually, this can be regarded as multivariate nonparametric regression problem $y_i=f(x_i)+\epsilon_i$ with $x_i$ unknown, and need to be found by optimization of the likelihood. The computational complexity of Gaussian process approach is cubic in the number of data points $n$, although approximation algorithm can be designed to reduce the complexity.

In this contribution, we propose a novel Bayesian approach to nonlinear PCA which puts priors on both $x$ and $V$. The model is based on an observation model similar to (\ref{eqn:obs}), but with two differences. First, the linear transformation is defined through the orthonormal matrix $V$ instead of $W$ which roughly corresponds to $VD$ in PPCA. Second, the linear transformation $V$ in our model is dependent on the corresponding latent variable. The linear transformations in different parts of the latent space are related by putting a Markov random field prior over the space of orthonormal matrices which makes the model identifiable. The model is estimated by Gibbs sampling which explores the posterior distribution of both the latent space and the transformation space. The computational burden for each iteration of Gibbs sampling is square in the number of data points.

The rest of the paper is organized as follows: In the next section, we present the Baysian model and discuss the Gibbs sampling estimation procedure. Since we think the readers might not be familiar with the von Mises-Fisher distribution, some background material is also provided. Some experiments are carried out in section 3 using both simulated manifold data and the handwritten digits data. We conclude in section 4 with some thoughts on possible extensions of the model. 

\section{Bayesian Nonlinear PCA}
\subsection{Stiefel Manifold and von Mises-Fisher Distribution}
Orthonormal matrices play a key role in our Bayesian model. By definition, the set of $n\times d$ matrices $X$ with $X^TX=I_{n\times n}$ is called the Stiefel manifold and denoted by $\nu_{n,d}$. This is a compact manifold. The most common probability distribution on the Stiefel manifold is the von Mises-Fisher distribution with a density with respect to the uniform distribution on the Stiefel manifold, which has an exponential family form:
\[p(X|C)\propto exp\{tr(C^TX)\}\]
where $C$ is a matrix of the same dimension as $X$ and the normalizing constant is omitted above. This distribution is denoted by $vMF(C)$. Note $vMF(0)$ is just the uniform distribution on the Stiefel manifold. 

Suppose the singular value decomposition of $C$ is $C=UDV^T$, with $U$ and $V$ being $n\times d$ and $d\times d$ orthonormal matrices, and $D$ a diagonal matrix containing the singular values of $C$. The density $p(X|C)$ is maximized at $X=UV^T$ which gives the ``most likely" matrix from the Stiefel manifold under this distribution. The diagonal matrix $D$ can be regarded as the concentration parameter of the distribution which determines the closeness of samples to the mode. Larger entries in $D$ makes the distribution more peaked around the mode $UV^T$. 

Sampling from von Mises-Fisher distribution has been studied in detail in \cite{hoffarxiv07}. Two efficient algorithms are proposed. One is the rejection sampling approach. The simplest proposal distribution for rejection sampling is the uniform distribution on the Stiefel manifold. Sampling randomly from $\nu_{n,d}$ can be done as follows \cite{hoff07}:
\begin{itemize}
\item Sample $u_1$ uniformly from the unit sphere $S_{n-1}$, and set $v_1=u_1$.
\item Sample $u_2$ uniformly from the unit sphere $S_{n-2}$ and set $v_2=N_1u_2$ where $N_1$ is an orthonormal matrix whose columns spanned the subspace orthogonal to $v_1$. 
\item[] $\;\;\vdots$
\item Sample $u_d$ uniformly from the unit sphere $S_{n-d}$ and set $v_d=N_du_d$ where $N_d$ is an orthonormal matrix whose columns span the subspace orthogonal to $v_1,\ldots,v_d$.
\end{itemize}
In \cite{hoffarxiv07}, more efficient rejection sampling is presented using a better proposal distribution.  Yet another approach in \cite{hoffarxiv07} is to use iterative Gibbs sampling on each column of $X$ based on the full conditional density. In our implementation, we use the rejection sampling approach, the R code of which is available from the website of the author of \cite{hoffarxiv07}. In \cite{hoff07}, von Mises-Fisher distribution aided with Gibbs sampling is used to build a Bayesian model for PCA.  Our model can also be regarded as a nonlinear extension of that work.

\subsection{Nonlinear PCA model with MRF}
The observation model of our Bayesian approach is similar to (\ref{eqn:obs}) but with the additional flexibility that the linear transformation is dependent on the latent variable:
\begin{equation}\label{eqn:bpca}
y_i=V_{x_i}x_i+\epsilon_i
\end{equation}
$V_{x_i}, i=1,2,\ldots n$ are constrained to be orthonormal and depends on the latent variable $x_i$. This is one difference with previous approaches in \cite{tippingpca99},\cite{tipping99},\cite{lawrence05}, where the transformation matrix $W$ roughly corresponds to principle directions properly scaled by the singular values of the data matrix, see (\ref{eqn:mle}). The prior on the noise is the same as before: $\epsilon_i\sim N(0,\sigma^2I_{n\times n})$. We use a conjugate prior $Gamma(\eta,\eta\tau^2/2)$ on the precision parameter $1/\sigma^2$ so that the expectation of $1/\sigma^2$ is $1/\tau^2$. The prior on $x_i$ is an isotropic Gaussian $x_i\stackrel{i.i.d.}{\sim}N(0,a^2I_{d\times d})$. Note we don't necessarily have $a=1$ here. The reason is that after putting the orthonormal constraint on $V_x$, the scale information of the data is shifted to the latent variable $x$. In our implementation, we set $a^2$ to be the sample variance of each covariate of the data points, and averaged over $p$ covariates. We find the result is insensitive to the choice of $a$ as long as $a$ is large enough. It is also as good to use the (improper) uniform prior for $x_i$. 

An important task is the specification of the prior for $V_{x_i}, i=1,2,\ldots n$. Independent prior obviously will not work here since the parameter $V_x$ typically has more degrees of freedom than can be estimated by the single constraining equation (\ref{eqn:bpca}). Therefore, we seek a prior that takes into account the correlation of transformation matrices for all $i$ simultaneously. A natural correlation between those orthonormal matrices can be introduced by the assumption that the transformation evolves slowly over the latent space. That is, the closeness of $x_i$ and $x_j$ for a pair $(x_i,x_j)$ as measured by the Euclidean distance in the latent space implies the closeness of $V_{x_i}$ and $V_{x_j}$ on the Stiefel manifold. 

Markov Random Field (MRF) is particularly useful for studying spatial models where the strength of interaction between random variables depends on the closeness of the corresponding sites. It has been widely used in image analysis and computer vision (e.g. \cite{li95},\cite{winkler03}). Formally, let $S$ be a finite index set representing the sites, with a random variable $Z_s$ associated with each site $s\in S$ and taking values in a subset of a Hilbert space with inner product $\langle\cdot,\cdot\rangle$. A neighborhood system is defined on the sites so that the full conditional probability of $Z_s$ only depends on its neighbors. One can think of the neighborhood system as an undirected graph where each vertex represents one site and two sites are neighbors of each other if and only if there is an edge connecting the two vertices. Although generally the conditional probabilities uniquely determines the joint distribution, the existence of the joint distribution is difficult to ascertain from the conditional ones. Thus it is generally more convenient to start by defining the joint distribution of the random variables. 

One simple example of MRF is defined by the joint distribution of all random variables: 
\[p(\{Z_s\})\propto exp\{\sum_{s\sim t}\lambda_{st}\langle Z_sZ_t\rangle\}\]
where the sum is over all pairs $(s,t)$ that are neighbors of each other. This distribution represent the pairwise interactions of random variables between neighbors. In our case, the sites are represented by the position of the latent variables $x_i$ in the latent space $R^d$. At each site, we attach a random variable $V_{x_i}$ taking values on the Stiefel manifold. The MRF prior for the $n$ orthonormal matrices $V_{x_i}, i=1,\ldots,n$ is defined by the joint density with respect to the uniform measure:
\[ p(\{V_{x_i}\}|\{x_i\})\propto exp\{tr(\sum\lambda_{ij}V_{x_i}^TV_{x_j})\}\]
where the sum is over all pairs of data points, i.e., the neighborhood system is defined by the complete graph that puts an edge between all pairs of sites. For ease of notation, this joint distribution is denoted by $MRF(\lambda_{ij})$. The scalar $\lambda_{ij}$ represents the strength of interaction between sites $i$ and $j$ and its choice is discussed later. Thus in our prior, the full conditional probability $p(V_{x_i}|V_{x_j},j\neq i)$ (omitting the conditioning on $x_i$ for simplicity) cannot be further reduced. The interaction between variables in this model is still additive in a pairwise manner though. 

The above probability density is well defined since the Stiefel manifold is compact and the normalizing constant can be found at least in theory. The conditional probability is trivially
\[p(V_{x_i}|V_{x_j},j\neq i)\propto exp\{tr( (\sum_{j,j\neq i}\lambda_{ij}V_{x_j})^T V_{x_i})\}\]
which is a von Mises-Fisher density with parameter $C=\sum_{j,j\neq i}\lambda_{ij}V_{x_j}$. 

As discussed previously, the mode of the conditional distribution $p(V_{x_i}|V_{x_j},j\neq i)$ can be found from the singular value decomposition of the matrix $\sum_{j,j\neq i}\lambda_{ij}V_{x_j}$. The decomposition is difficult to find in closed form, but some approximation can give some insight into the prior. Suppose that $\lambda_{ij}$ is large when $x_i$ and $x_j$ are close and negligible when they are distant from each other. Besides, if for those $x_j$ close to $x_i$, the corresponding matrices $V_{x_j}$ are also close to each other and approximated by a common orthonormal matrix $V$, then $\sum_{j,j\neq i}\lambda_{ij}V_{x_j}$ can be approximated by $(\sum_{j\neq i}\lambda_{ij})V$. The mode of the distribution is approximately $V$ and $\sum_{j\neq i}\lambda_{ij}$ determines the concentration of the distribution. So the effect of the MRF prior is to smooth the transformation matrices $V_{x_i}$ so that sites close by in the latent space are associated with similar transformations.

By the above discussion, we would like to specify $\lambda_{ij}$  as a decreasing function of the Euclidean distance between $x_i$ and $x_j$, $||x_i-x_j||$. In this work, we make use of a Gaussian kernel function for this purpose:
\[\lambda_{ij}=c\phi(||x_i-x_j||/w) \]
where $\phi(x)=exp\{-x^2/2\}$. The kernel width $w$ determines the relative influence of different  sites and the parameter $c$ is related to the concentration of the conditional distribution and thus affects the ``smoothness" of the joint distribution of $\{V_{x_i}\}$.

Summarizing, we use the following model for nonlinear dimension reduction:
\begin{eqnarray*}
y_i&=& V_{x_i}x_i+\epsilon_i\\
\{V_{x_i}\}|\{x_i\}&\sim &MRF(\{\lambda_{ij}\}),\lambda_{ij}=c\phi(||x_i-x_j||/w)\\
x_i&\sim& N(0,a^2I)\\
\epsilon_i|\sigma^2&\sim& N(0,\sigma^2I)\\
\frac{1}{\sigma^2}&\sim& Gamma(\eta,\eta\tau^2/2)
\end{eqnarray*}
% needed in second column of first page if using \pubid
%\pubidadjcol
We choose $a^2$ to be a large number or even infinity. Similar to \cite{hoff07}, we set the ``prior sample size" $\eta=2$, and $\tau^2$ is derived from a pilot dimension reduction study such as the traditional PCA. For example, we can use $\tau^2=\sum_i||y_i-\hat{y}_i||^2/np$, where $\hat{y}_i$ is the reconstructed data point from $d-$dimensional PCA. The choice of $c$ and $w$ is more difficult. For full Bayesian analysis, we should put a prior on $c$ and $w$ also. But this will cause computational difficulty with Gibbs sampling. In our experience, the choice $w=\sum_{i<j} ||x_i-x_j||/{n\choose 2}$ and $c=100/n$ generally gives satisfactory results. 
\subsection{Posterior Computation}
We propose using Gibbs sampling for posterior computation. The full conditional distribution of $V_{x_i}$ is 
\begin{eqnarray*}
&&p(V_{x_i}|V_{x_j},j\neq i, \{x_k\}_{k=1}^n, \sigma^2, Y)\\
&\propto &exp\{-\frac{(y_i-V_{x_i}x_i)^T(y_i-V_{x_i}x_i)}{2\sigma^2}\}\cdot
exp\{tr(\sum_{j\neq i}\lambda_{ij}V_{x_i}^TV_{x_j})\}\\
&\propto&exp\{tr(V_{x_i}^T [y_ix_i^T/\sigma^2+\sum_{j\neq i}\lambda_{ij}V_{x_j}]\}
\end{eqnarray*}
The expressions for other full conditional distributions are standard and their derivations omitted. The Gibbs sampling then iterates between the following steps.
\begin{itemize}
\item update $V_{x_i}$, for $i=1,\ldots, n$, by sampling from $vMF(C)$ with $C=y_ix_i^T/\sigma^2+\sum_{j\neq i}\lambda_{ij}V_{x_j}$.
\item update the latent variables $x_i$, for $i=1,\ldots, n$, by sampling from 
\[x_i|V_{x_i},y_i,\sigma^2\sim N(\frac{a^2}{a^2+\sigma^2}V_{x_i}^Ty_i, \frac{a^2\sigma^2}{a^2+\sigma^2}).\]
\item update the parameter $\sigma^2$ by sampling $1/\sigma^2$ from $Gamma((\eta+np)/2, (\eta\tau^2+\sum_i||y_i-V_{x_i}x_i||^2)/2)$.
\end{itemize}
The  Gibbs sampling algorithm is initialized using standard PCA, setting the parameters and variable to the corresponding variables obtained from singular value decomposition of the data matrix. For statistical inferences of the parameters, the most convenient approach is to use the posterior sample average after the ``burn in" period.
% Reminder: the "draftcls", not "draft", class option should be used if
% it is desired that the figures are to be displayed while in draft mode.

\section{Experimental Results}
In this section, we perform some limited experiments to illustrate our nonlinear Bayesian model for dimension reduction. 

\begin{figure*}
\centerline{
\subfigure[]{\includegraphics[width=2.5in]{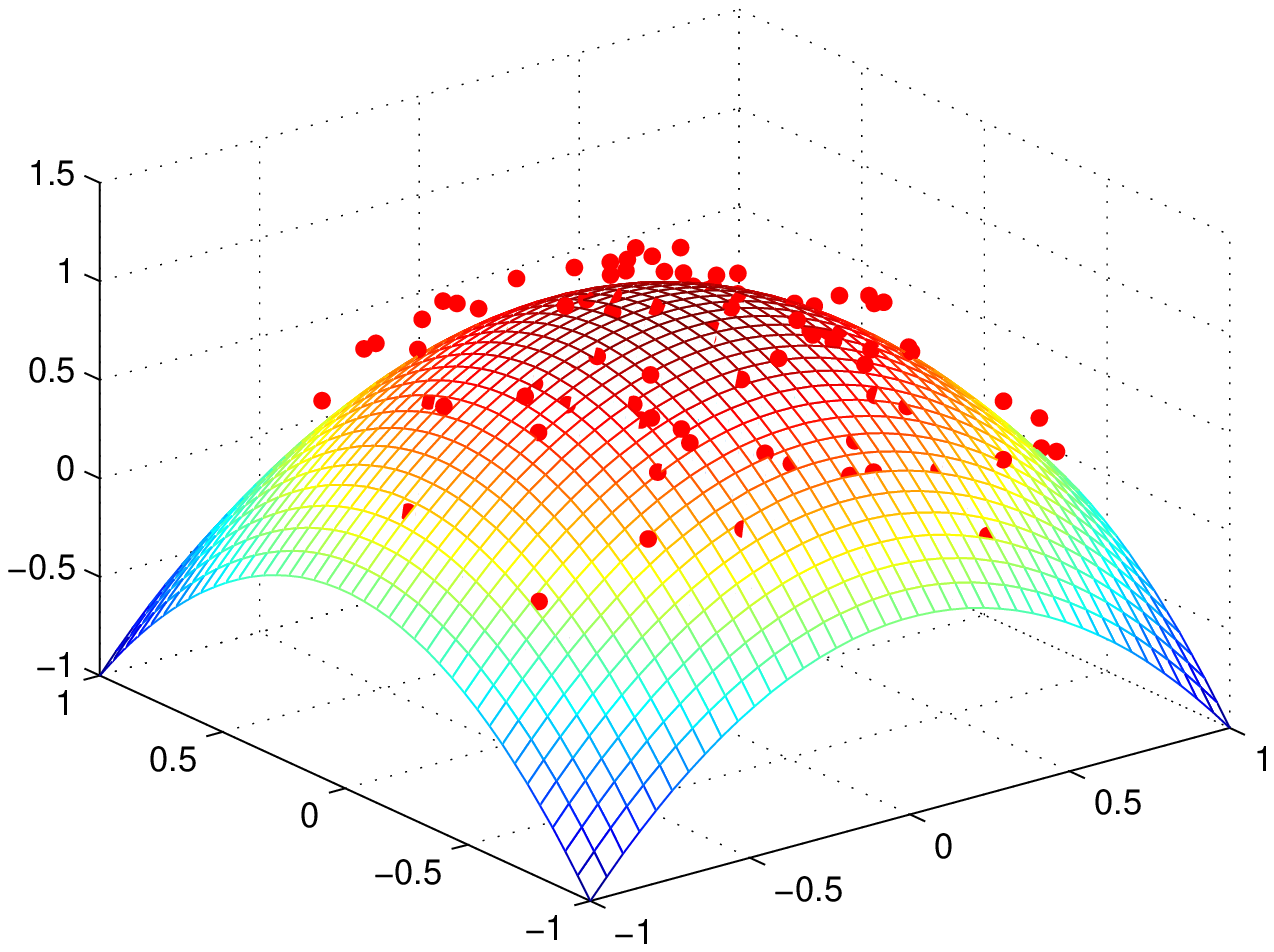}
%\label{fig_second_case}
}
\hfil
\subfigure[]{\includegraphics[width=2.5in]{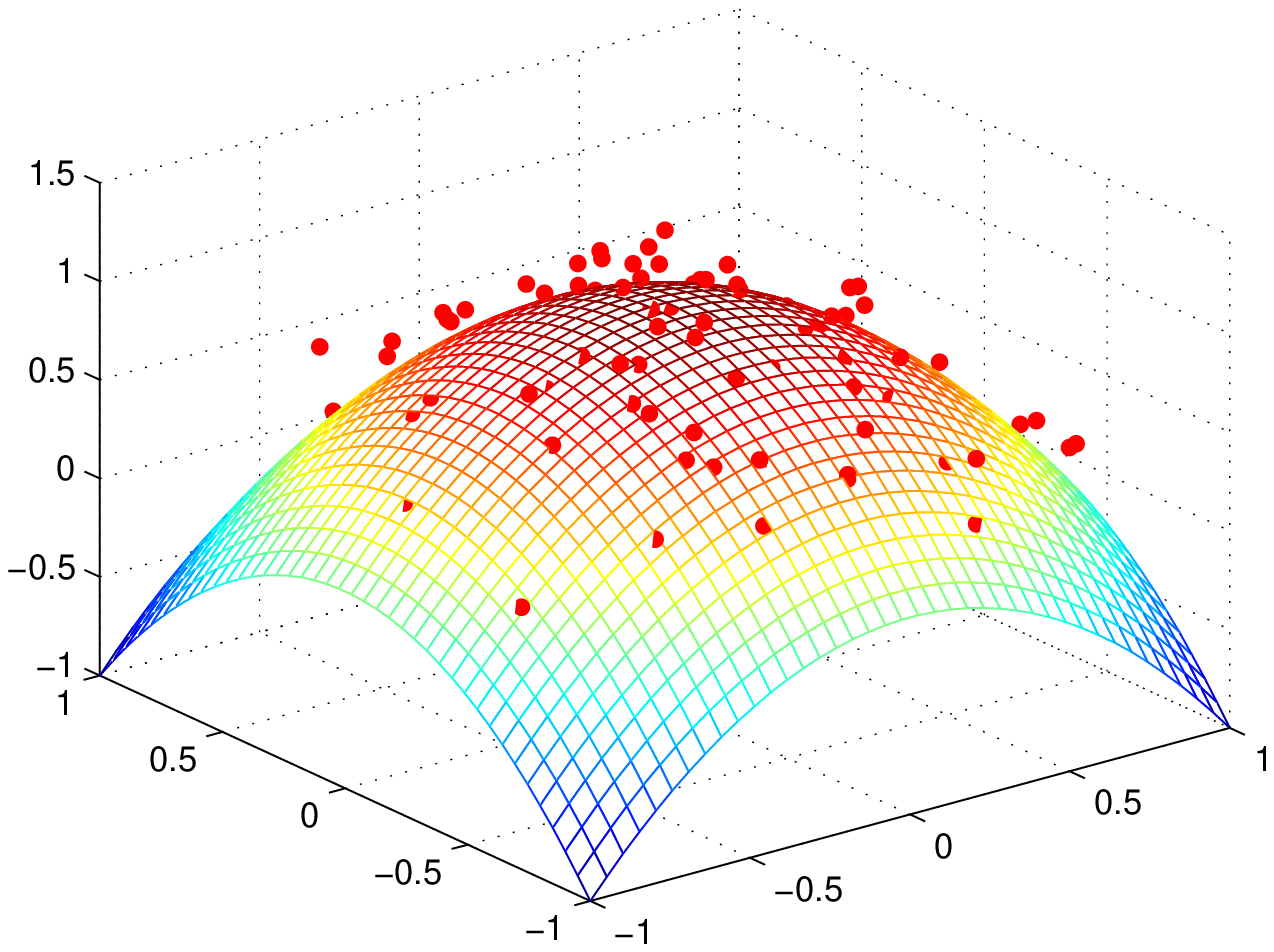}
%\label{fig_second_case}
}
}
\centerline{\subfigure[]{\includegraphics[width=2.5in]{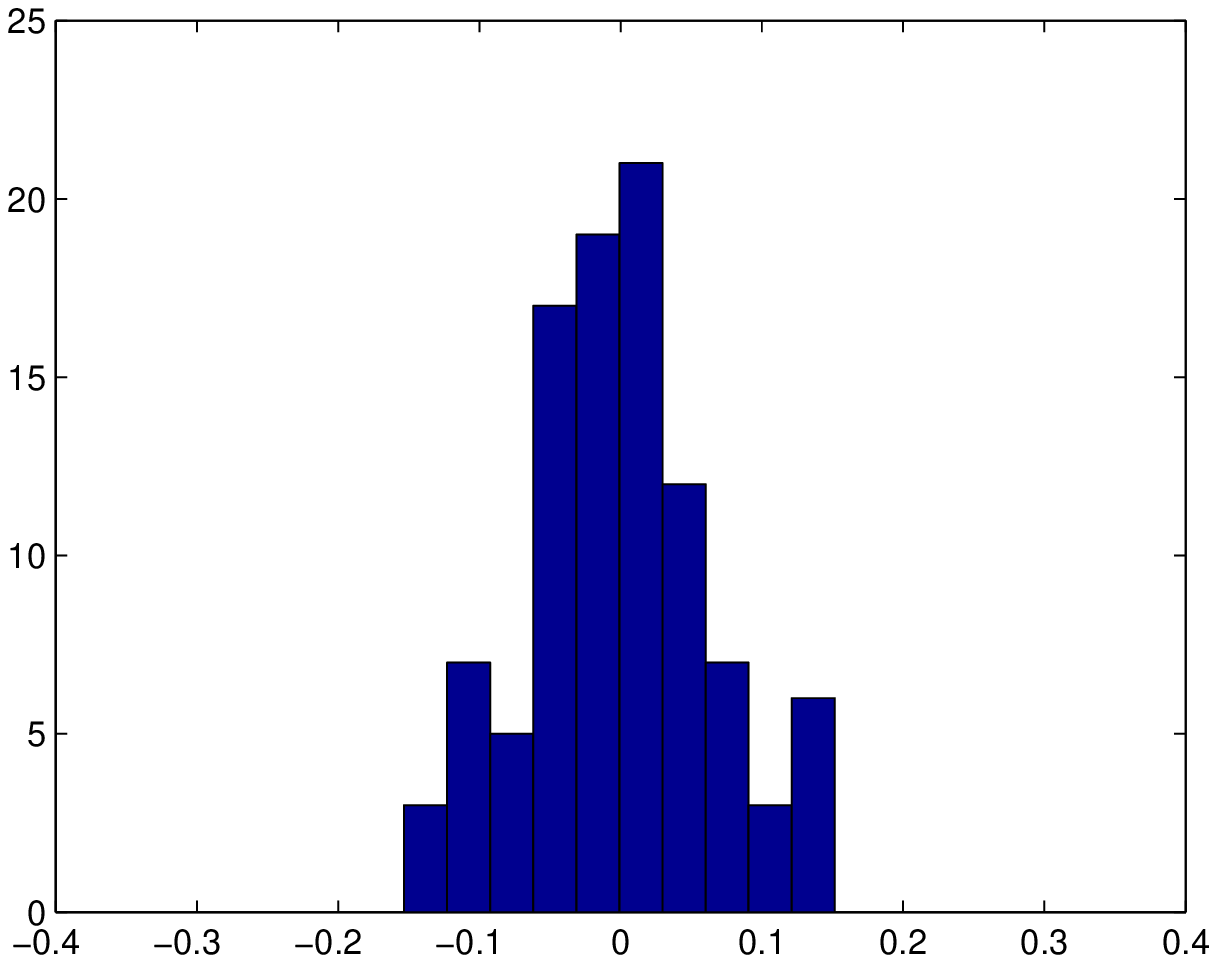}
%\label{fig_first_case}
}
\hfil
\subfigure[]{\includegraphics[width=2.5in]{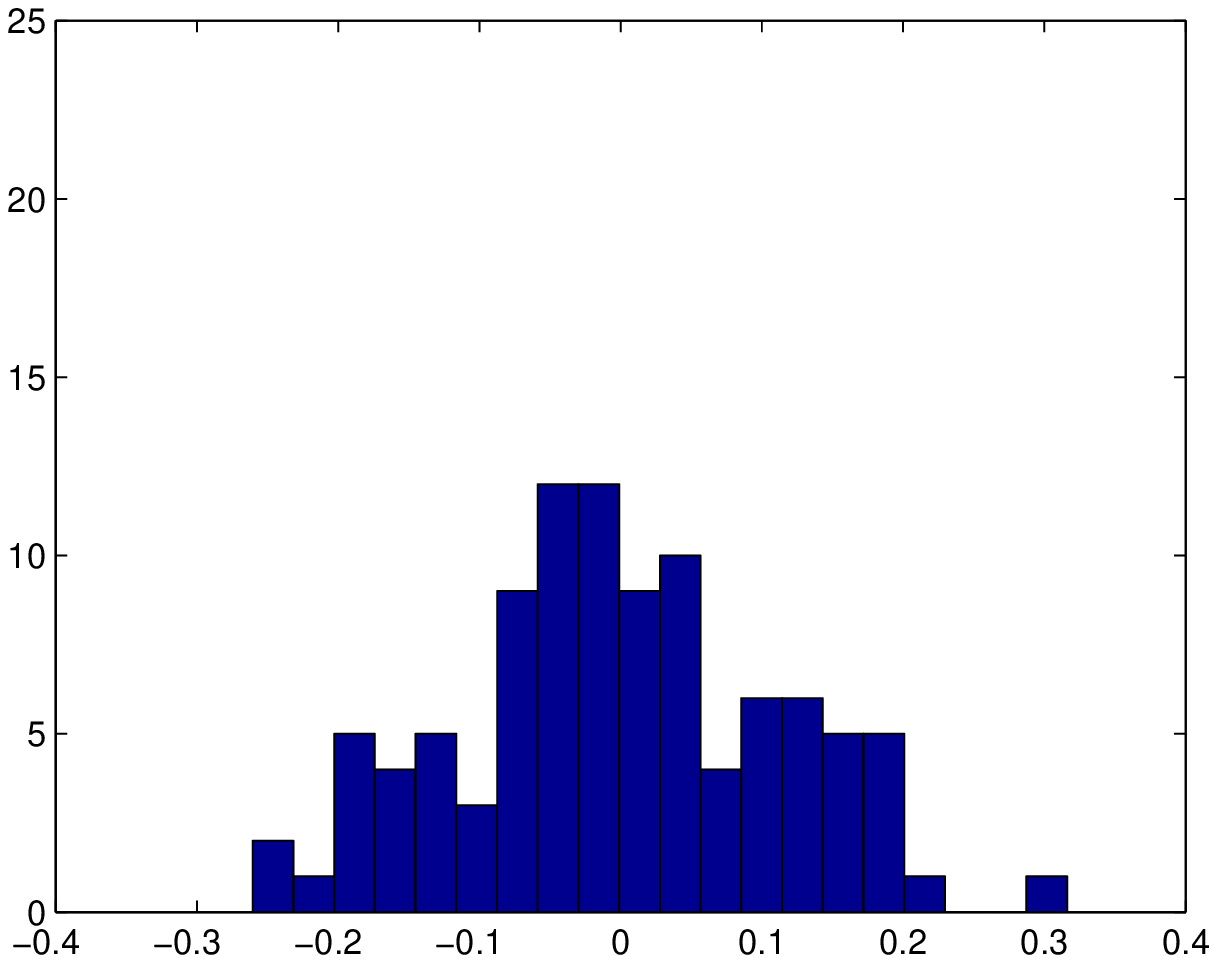}
%\label{fig_second_case}
}
\hfil
\subfigure[]{\includegraphics[width=2.5in]{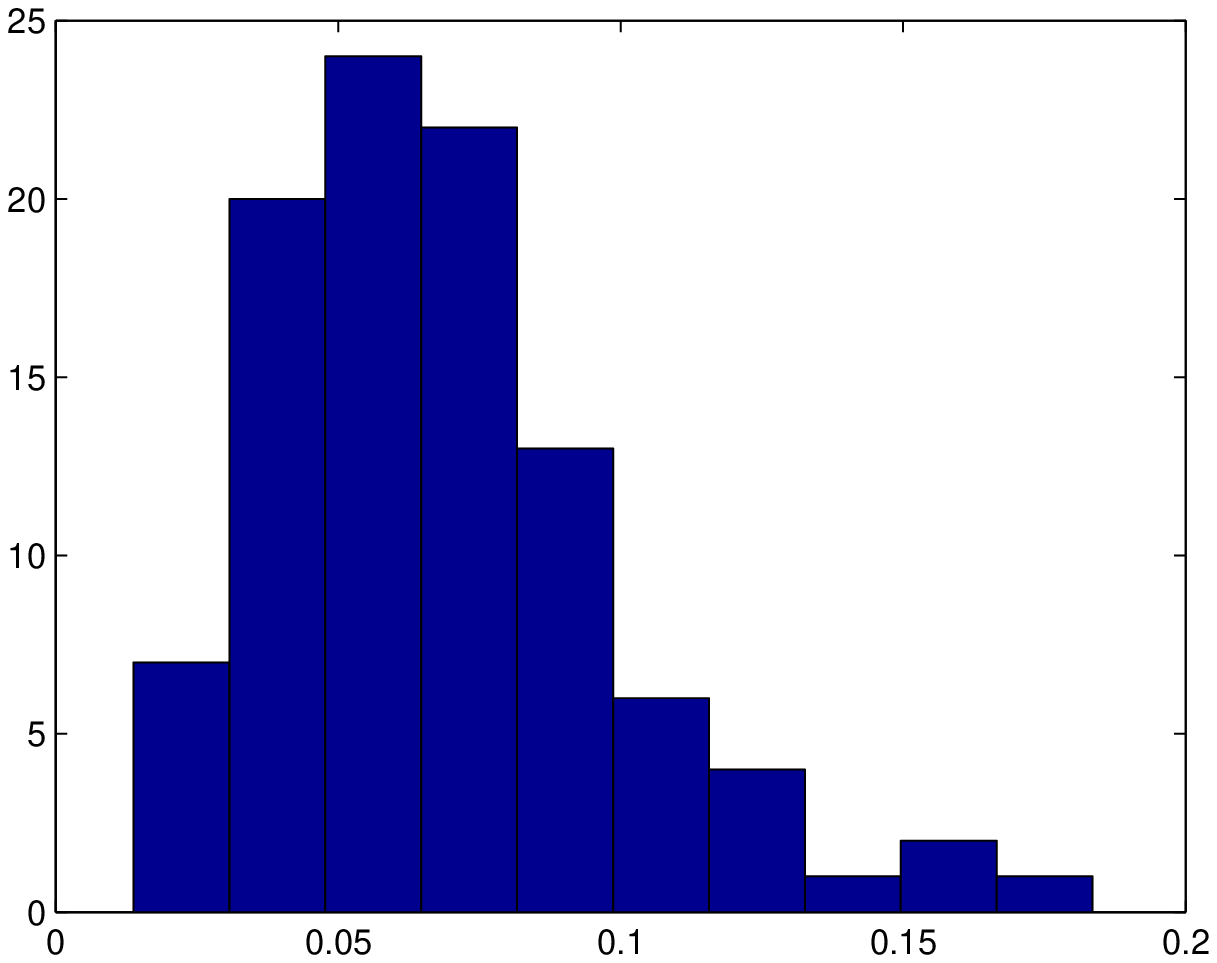}
%\label{fig_second_case}
}
}
\caption{(a) Simulated $100$ points with added noise on the surface of a unit sphere. (b) Reconstructed $100$ points. (c) Histogram of distances of $100$ simulated points to the sphere. (d) Histogram of distances of reconstructed points to the sphere. (e) Histogram of reconstruction errors.}
\label{fig:sphere}
\end{figure*}

To demonstrate the nonlinearity of the model, we sample $100$ points on the unit sphere with noise level $\sigma=0.05$. The data is shown on Fig \ref{fig:sphere}. The Bayesian model is fitted with latent space dimension $d=2$. The reconstructed data points from the latent space representation is also shown on Fig. \ref{fig:sphere}. We can compare the histograms of the distances of data points to the surface. We also show the histogram of the reconstruction errors to illustrate the accuracy of the reconstruction. 

\begin{figure*}
\centerline{\subfigure[]{\includegraphics[width=2.5in]{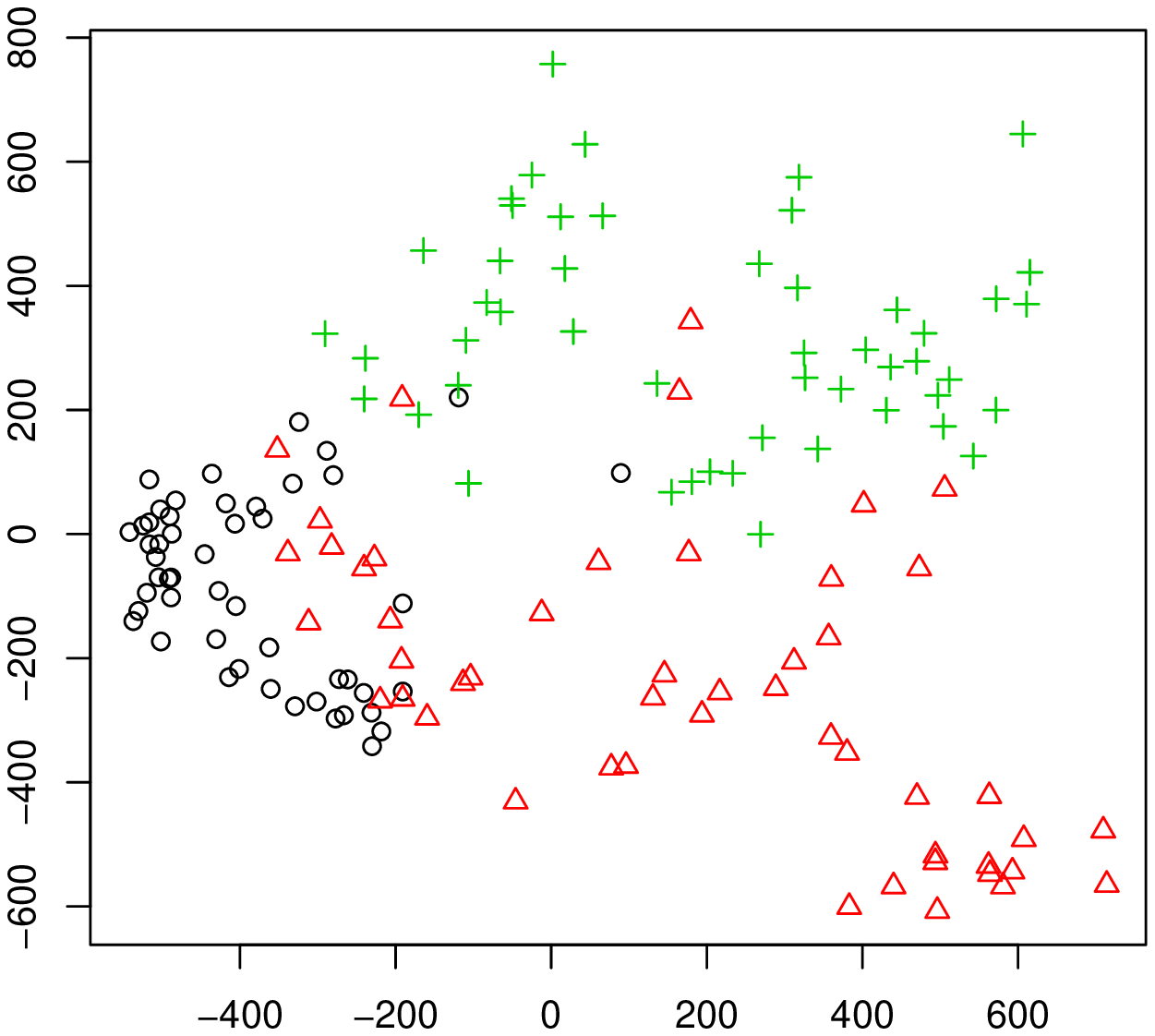}
%\label{fig_first_case}
}
\hfil
\subfigure[]{\includegraphics[width=2.5in]{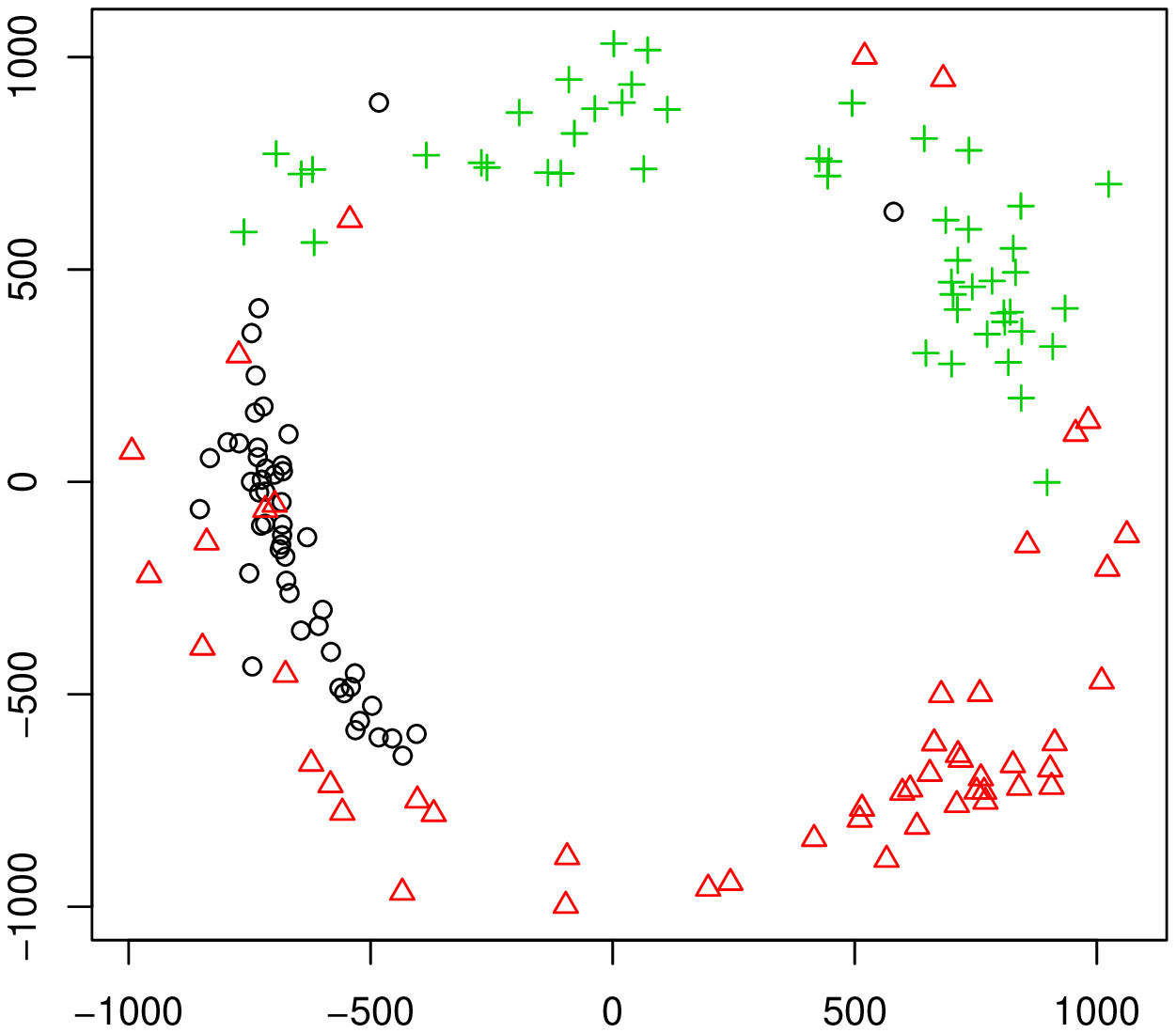}
%\label{fig_second_case}
}
}
\caption{Visualization results for the handwritten digits data. (a) Projection given by PCA. (b) Projection given by our model. `1' is represented by circles, `2' by triangles, and `3' by pluses.}
\label{fig:digits}
\end{figure*}

One popular dataset for visualization of dimension reduction is handwritten digits. We therefore use the MNIST database of handwritten digits and apply the model to a subset of 150 digits 1, 2, 3 (50 of each). The image dimension of this dataset is $28\times 28$. To reduce computational complexity, we subsampled the images so that the dimension is reduced to $14\times 14$. The position of each image in the latent space is shown in Fig. \ref{fig:digits}, together with that obtained by traditional PCA. An objective assessment can be obtained by counting the number of images whose nearest neighbor in the latent space represents a different digit. For traditional PCA, we have 53 such images, while we only have 25 such images in our new model. 

% An example of a floating figure using the graphicx package.
% Note that \label must occur AFTER (or within) \caption.
% For figures, \caption should occur after the \includegraphics.
%
%\begin{figure}
%\centering
%\includegraphics[width=2.5in]{myfigure.eps}
%\caption{Simulation Results}
%\label{fig_sim}
%\end{figure}

% An example of a double column floating figure using two subfigures.
% (The subfigure.sty package must be loaded for this to work.)
% The subfigure \label commands are set within each subfigure command, the
% \label for the overall figure must come after \caption.
% \hfil must be used as a separator to get equal spacing
%

% An example of a floating table. Note that, for IEEE style tables, the
% \caption command should come BEFORE the table. Table text will default to
% \footnotesize as IEEE normally uses this smaller font for tables.
% The \label must come after \caption as always.
%
%\begin{table}
%% increase table row spacing, adjust to taste
%\renewcommand{\arraystretch}{1.3}
%\caption{An Example of a Table}
%\label{table_example}
%\centering
%% The array package and the MDW tools package offers better commands
%% for making tables than plain LaTeX2e's tabular which is used here.
%\begin{tabular}{|c||c|}
%\hline
%One & Two\\
%\hline
%Three & Four\\
%\hline
%\end{tabular}
%\end{table}

\section{Conclusion}
We have presented a novel Bayesian framework for performing nonlinear principal component analysis. Each data point is associated with a different transformation and the different transformations are smoothed by a MRF type prior. We demonstrated with some experiments that our new model can discover nonlinear structure underlying the datasets. 

As in traditional PCA, dimension selection is a difficult problem in our problem. We are currently investigating the possibility of automatic dimension selection as done in \cite{hoff07} by putting a prior on the dimension. This seems to be a viable approach. 

Although the computational complexity for our model is square in the number of samples, which compares favorably with the approach adopted in \cite{lawrence05}. It is still desirable to reduce the computation if possible. The MRF prior used in our current implementation corresponds to a complete graph. It is possible to use a sparser graph that only connects nearby points in the latent space. This strategy can further reduce the computational complexity. 
% or

% Use this command to get the appendices' numbers in "A", "B" instead of the
% default capitalized Roman numerals ("I", "II", etc.).
% However, the capital letter form may result in awkward subsection numbers
% (such as "A-A"). Capitalized Roman numerals are the default.
%\useRomanappendicesfalse
%
%\appendices

%\section{}

\section*{Acknowledgment}
% optional entry into table of contents (if used)
%\addcontentsline{toc}{section}{Acknowledgment}
This work was supported by MOE Tier 1 SUG administered by Nanyang Technological University.



% http://www.ctan.org/tex-archive/macros/latex/contrib/supported/IEEEtran/
%\bibliographystyle{IEEEtran.bst}

\bibliographystyle{IEEEtran}
\bibliography{arxiv}

\end{document}